\documentclass[9pt]{article} 
\usepackage{nips15submit_e,times}
\usepackage{hyperref}
\usepackage{url}

\usepackage{amsmath,amssymb,amsfonts}
\usepackage{url}
\usepackage{cite}
\usepackage{graphicx}
\usepackage{multirow}

\def\RR{\mathbb R}

\graphicspath{{figures/}}

\nipsfinalcopy

\title{Deep clustering: Discriminative embeddings for segmentation and separation}

\author{
	John R. Hershey\\
	MERL\\
	Cambridge, MA\\
	\texttt{hershey@merl.com} \\
	\And
	Zhuo Chen\\
	Columbia University\\
	New York, NY\\
	\texttt{zc2204@columbia.edu} \\
	\AND
	Jonathan Le Roux\\
	MERL\\
	Cambridge, MA\\
	\texttt{leroux@merl.com} \\
	\And
	Shinji Watanabe\\
	MERL\\
	Cambridge, MA\\
	\texttt{watanabe@merl.com} \\
}

\begin{document}

\maketitle

\begin{abstract} 
We address the problem of acoustic source separation in a deep learning framework we call ``deep clustering''.   Rather than directly estimating signals or masking functions, we train a deep network to produce spectrogram embeddings that are discriminative for partition labels given in training data.   Previous deep network approaches provide great advantages in terms of learning power and speed, but previously it has been unclear how to use them to  separate signals in a class-independent way. In contrast, spectral clustering approaches are flexible with respect to the classes and number of items to be segmented, but it has been unclear how to leverage the learning power and speed of deep networks. To obtain the best of both worlds, we use an objective function that to train embeddings that yield a low-rank approximation to an ideal pairwise affinity matrix, in a class-independent way.   This avoids the high cost of spectral factorization and instead produces compact clusters that are amenable to simple clustering methods.   The segmentations are therefore implicitly encoded in the embeddings, and can be "decoded" by clustering. Preliminary experiments show that the proposed method can separate speech: when trained on spectrogram features containing mixtures of two speakers, and tested on mixtures of a held-out set of speakers, it can infer masking functions that improve signal quality by around 6dB.  We show that the model can generalize to three-speaker mixtures despite training only on two-speaker mixtures.   The framework can be used without class labels, and therefore has the potential to be trained on a diverse set of sound types, and to generalize to novel sources.    We hope that future work will lead to segmentation of arbitrary sounds, with extensions to microphone array methods as well as image segmentation and other domains.
\end{abstract} 
\section{Introduction}
\label{sec:intro}

In real world perception, we are often confronted with the problem of selectively attending to objects whose features are intermingled with one another in the incoming sensory signal.   In computer vision, the problem of scene analysis is to partition an image or video into regions attributed to the visible objects present in the scene.   In audio there is a corresponding problem known as auditory scene analysis\cite{bregman1994auditory,Darwin95}, which seeks to identify the components of audio signals corresponding to individual sound sources in a mixture signal.   Both of these problems can be approached as \emph{segmentation} problems, where we formulate a set of ``\emph{elements}'' in the signal via an indexed set of features, each of which carries (typically multi-dimensional) information about part of the signal.  For images, these elements are typically defined spatially in terms of pixels, whereas for audio signals they may be defined in terms of time-frequency coordinates.    The segmentation problem is then solved by segmenting elements into groups or partitions, for example by assigning a group label to each element.  Note that although \emph{clustering} methods can be applied to segmentation problems, the segmentation problem is technically different in that clustering is classically formulated as a domain-independent problem based on simple objective functions defined on pairwise point relations, whereas partitioning may depend on complex processing of the whole input, and the task objective may be arbitrarily defined via training examples with given segment labels.  

Segmentation problems can be broadly categorized into \emph{class-based} segmentation problems where the goal is to learn from training class labels to label known object classes, versus more general \emph{partition-based} segmentation problems where the task is to learn from labels of partitions, without requiring object class labels, to segment the input.   Solving the partition-based problem has the advantage that unknown objects could then be segmented.  In this paper, we propose a new partition-based approach which learns \emph{embeddings} for each input elements, such that the correct labeling can be determined by simple clustering methods.   We focus on the single-channel audio domain, although our methods are applicable to other domains such as images and multi-channel audio.  The motivation for segmenting in this domain, as we shall describe later, is that using the segmentation as a mask, we can extract parts of the target signals that are not corrupted by other signals. 

Since class-based approaches are relatively straightforward, and have been tremendously successful at their task, we first briefly mention this general approach.   In class based vision models, such as \cite{farabet2013learning,sharma2014recursive,smaragdis2007convolutive}, a hierarchical classification scheme is trained to estimate the class label of each pixel or super-pixel region.  In the audio domain, single-channel speech separation methods, for example, segment the time-frequency elements of the spectrogram into regions dominated by a target speaker, either based on classifiers \cite{weiss2006estimating,kim2009algorithm,wang2013exploring}, or generative models\cite{Roweis03,schmidt2006single, hershey2010super}.   In recent years, the success of deep neural networks for classification problems has naturally inspired their use in class-based segmentation problems \cite{sharma2014recursive,wang2013towards}, where they have proven very successful.  

However class-based approaches have some important limitations.  First, of course, the assumed task of labeling known classes fundamentally does not address the general problem in real world signals that there may be a large number of possible classes, and many objects may not have a well-defined class.   It is also not clear how to directly apply current class-based approaches to the more general problem.   Class-based deep network models for separating sources require explicitly representing output classes and object instances in the output nodes, which leads to complexities in the general case.  Although generative model-based methods can in theory be flexible with respect to the number of model types and instances after training, inference typically cannot scale computationally to the potentially larger problem posed by more general segmentation tasks.

In contrast,  humans seem to solve the partition-based problem, since they can apparently segment well even with novel objects and sounds.  This observation is the basis of Gestalt theories of perception, which attempt to explain perceptual grouping in terms of features such as proximity and similarity\cite{wertheimer1938laws}.  The partition-based segmentation task is closely related, and follows from a tradition of work in image segmentation and audio separation.  Application of the perceptual grouping theory to audio segmentation is generally known as computational auditory scene analysis (CASA) \cite{Cooke91,Ellis96}.    

\emph{Spectral clustering} is an active area of machine learning research with application to both image and audio segmentation.    It uses local affinity measures between features of elements of the signal,  and optimizes various objective functions using spectral decomposition of the normalized affinity matrix \cite{shi2000normalized}.   In contrast to conventional \emph{central clustering} algorithms such as $k$-means, spectral clustering has the advantage that it does not require points to be tightly clustered around a central prototype, and can find clusters of arbitrary topology, provided that they form a connected sub-graph.  Because of the local form of the pairwise kernel functions used,  in difficult spectral clustering problems the affinity matrix has a sparse block-diagonal structure that is not directly amenable to central clustering, which works well when the block diagonal affinity structure is dense.  The powerful but computationally expensive eigenspace transformation step of spectral clustering addresses this, in effect, by "fattening" the block structure, so that connected components become dense blocks, prior to central clustering \cite{bach2006learning}.  

Although affinity-based methods were originally unsupervised inference methods, multiple-kernel learning methods such as \cite{fowlkes2003learning, bach2006learning} were later introduced to train weights used to combine separate affinity measures.  This allows us to consider using them for partition-based segmentation tasks in which partition labels are available, but  without requiring specific class labels.     In \cite{bach2006learning}, this was applied to speech separation by including a variety of complex features developed to implement various auditory scene analysis grouping principles, such as similarity of onset/offset, pitch, spectral envelope, and so on, as affinities between time-frequency regions of the spectrogram.  The input features included a dual pitch-tracking model in order to improve upon the relative simplicity of kernel-based features, at the expense of generality.

Rather than using specially designed features and relying on the strength of the spectral clustering framework to find clusters,  we propose to use deep learning to derive embedding features that make the segmentation problem amenable to simple and computationally efficient clustering algorithms such as $k$-means, using the partition-based training approach. 
Learned feature transformations known as \emph{embeddings} have recently been gaining significant interest in many fields.      Unsupervised embeddings obtained by auto-associative deep networks, used with relatively simple clustering algorithms,  have recently been shown to outperform spectral clustering methods \cite{tian2014learning, huang2014deep} in some cases.    Embeddings trained using pairwise metric learning,  such as word2vec \cite{mikolov2013distributed} using neighborhood-based partition labels,  have also been shown to have interesting invariance properties.  We present below an objective function that minimizes the distances between embeddings of elements within a partition, while maximizing the distances between embeddings for elements in different partitions.   This appears to be an appropriate criterion for central clustering methods.   The proposed embedding approach has the attractive property that all partitions and their permutations can be represented implicitly using the fixed-dimensional output of the network.

The experiments described below show that the proposed method can separate speech using a speaker-independent model with an open set of speakers at test time.
As in \cite{bach2006learning}, we derive partition labels by mixing signals together and observing their spectral dominance patterns.  After training on a database of mixtures of speakers trained in this way, we show that without any modification the model shows a promising ability to separate  three-speaker mixtures despite training only on two-speaker mixtures.  Although results are preliminary, the hope is that this work leads to methods that can achieve class-independent segmentation of arbitrary sounds, with additional application to image segmentation and other domains.

\section{Learning deep embeddings for clustering}
\label{sec:model}

We define as $x$ a raw input signal, such as an image or a time-domain waveform, and as $X_n = g_n(x), n\in \{1,\dots,N\},$ a feature vector indexed by an element $n$. In the case of images, $n$ typically may be a superpixel index and $X_n$ some vector-valued features of that superpixel; in the case of audio signals, $n$ may be a time-frequency index $(t,f)$, where $t$ indexes frames of the signal and $f$ frequencies, and $X_n = X_{t,f}$ the value of the complex spectrogram at the corresponding time-frequency bin.
We assume that there exists a reasonable partition of the elements $n$ into regions, which we would like to find, for example to further process the features $X_n$ separately for each region. In the case of audio source separation, for example, these regions could be defined as the sets of time-frequency bins in which each source dominates, and estimating such a partition would enable us to build time-frequency masks to be applied to $X_n$, leading to time-frequency representations that can be inverted to obtain isolated sources.

To estimate the partition, we seek a $K$-dimensional embedding $V = f_{\theta}(x) \in \RR^{N \times K}$, parameterized by $\theta$, such that performing some simple clustering in the embedding space will likely lead to a partition of $\{1,\dots,N\}$ that is close to the target one. 
In this work, $V = f_{\theta}(x)$ is based on a deep neural network that is a global function of the entire input signal $x$ (we allow for a feature extraction step to create the network input; in general, the input features may be completely different from $X_n$).   Thus our transformation can take into account global properties of the input, and the embedding can be considered a permutation- and cardinality-independent encoding of the network's estimate of the signal partition. 
Here we consider a unit-norm embedding, so that $|v_n|^2=\sum_{k} v_{n,k}^2 = 1, \forall n$, where $v_n=\{v_{n,k}\}$ and $v_{n,k}$ is the value of the $k$-th dimension of the embedding for element $n$. We omit the dependency of $V$ on $\theta$ to simplify notations.

The partition-based training requires a reference label indicator $Y = \{y_{n,c}\}$,  mapping each element $n$ to each of $c$ arbitrary partition classes, so that $y_{n,c} = 1$ if element $n$ is in partition $c$.   
For a training objective, we seek embeddings that enable accurate clustering according to the partition labels.
To do this, we need a convenient expression that is invariant to the number and permutations of the partition labels from one training example to the next. 
One such objective for minimization is 
\begin{align}
    C(\theta) &= |  VV^{T} - YY^{T} |_{W}^2     = \sum_{i,j : y_i = y_j}  \frac{(\langle v_i,v_j\rangle - 1)^2}{d_i} + \sum_{i,j : y_i \neq y_j}  \frac{(\langle v_i,v_j\rangle  - 0)^2}{\sqrt{d_i d_j}},
    \\ &= \sum_{i,j : y_i = y_j}  \frac{|v_i - v_j|^2}{d_i}    +  \sum_{i,j}  \frac{\left(|v_i - v_j|^2 - 2\right)^2}{4\sqrt{d_i d_j}} - N, \label{eq:obj_like_kmeans}
\end{align}
where $|A|_{W}^2 = \sum_{i,j} w_{i,j} a_{i,j}^2 $ is a weighted Frobenius norm, with $W = d^{-\frac{1}{2}}d^{-\frac{T}{2}}$, where $d_i = Y Y^T 1 $ is an $(N\times 1) $ vector of partition sizes: that is, $d_i = |\{j: y_{i}=y_{j}\} |$.   In the above we use the fact that $|v_n|^2=1,\forall n$.  
Intuitively, this objective pushes the inner product $\langle v_i,v_j\rangle$ to 1 when $i$ and $j$ are in the same partition, and to $0$ when they are in different partitions. Alternately, we see from (\ref{eq:obj_like_kmeans}) that it pulls the squared distance $|v_i - v_j|^2$ to 0 for elements within the same partition, while preventing the embeddings from trivially collapsing into the same point.  Note that the first term is the objective function minimized by $k$-means, as a function of cluster assignments, and in this context the second term is a constant.  So the objective reasonably tries to lower the $k$-means score for the labeled cluster assignments at training time.

This formulation can be related to spectral clustering as follows. We can define an ideal affinity matrix $A^* = YY^T$, that is block diagonal up to permutation and use an inner-product kernel, so that $A=VV^{T}$ is our affinity matrix. Our objective becomes $C =  |A - A^{*}|_{\mathrm{F}}^2$, which measures the deviation of the model's affinity matrix from the ideal affinity.
Note that although this function ostensibly sums over all pairs of data points $i,j$, the low-rank nature of the objective leads to an efficient implementation, defining $D=\operatorname{diag}( Y Y^T 1)$: 
\begin{align}
C &=  |VV^{T} - YY^{T}|_{\mathrm{W}}^2  = |V^{T}D^{-\frac{1}{2}}V|_{\mathrm{F}}^2 - 2 |V^{T}D^{-\frac{1}{2}}Y |_{\mathrm{F}}^2 + |Y^{T}D^{-\frac{1}{2}}Y|_{\mathrm{F}}^2,
\end{align}
which avoids explicitly constructing the $N \times N$ affinity matrix.  In practice, $N$ is orders of magnitude greater than $K$, leading to a significant speedup.
To optimize a deep network, we typically need to use first-order methods.   Fortunately derivatives of our objective function with respect to $V$ are also efficiently obtained due to the low-rank structure: 
\begin{align}
\frac{\partial C}{\partial V^T} &=   4 D^{-\frac{1}{2}}V V^{T}D^{-\frac{1}{2}}V- 4 D^{-\frac{1}{2}} Y Y^{T} D^{-\frac{1}{2}} V.
\end{align}

This low-rank formulation also relates to spectral clustering in that the latter typically requires the Nystr\"{o}m low-rank approximation to the affinity matrix, \cite{fowlkes2004spectral} for efficiency,  so that the singular value decomposition (SVD) of an $N \times K$ matrix can be substituted for the much more expensive eigenvalue decomposition of the $K \times K$ normalized affinity matrix.  Rather than following spectral clustering in making a low-rank approximation of a full-rank model, our method can be thought of as directly optimizing a low-rank affinity matrix so that processing is more efficient and parameters are tuned to the low-rank structure.

At test time, we compute the embeddings $V$ on the test signal, and cluster the rows $v_i \in \RR^K$, for example using $k$-means.  We also alternately perform a spectral-clustering style dimensionality reduction before clustering, starting with a singular value decomposition (SVD),  $\tilde{V} = U S R^{T}$, of normalized $\tilde{V} = D^{-\frac{1}{2}} V$, where  $D=VV^{T}1_{N}$, sorted by decreasing eigenvalue, and clustering the normalized rows of the matrix of $m$ principal left singular vectors, with the $i$'th row given by $\tilde{u}_{i,r} = u_{i,r} / \sqrt{\sum_{r'=1}^{m} u_{i,r'}} \; : r \in [1,m]$, similar to \cite{ng2002spectral}.

\section{Speech separation experiments}
\label{sec:exp}
\subsection{Experimental setup}
We evaluate the proposed model on a speech separation task: the goal is to separate each speech signal from a mixture of multiple speakers. While separating speech from non-stationary noise is in general considered to be a difficult problem, separating speech from  other speech signals is particularly challenging because all sources belong to the same class, and share similar characteristics.  Mixtures involving speech from same gender speakers are the most difficult since the pitch of the voice is in the same range. We here consider mixtures of two speakers and three speakers (the latter always containing at least two speakers of the same gender).  
However, our method is not limited in the number of sources it can handle or the vocabulary and discourse style of the speakers. 
To investigate the effectiveness of our proposed model, we built a new dataset of speech mixtures based on the Wall Street Journal (WSJ0) corpus, leading to a more challenging task than in existing datasets.
Existing datasets are too limited for evaluation of our model because, for example, the speech separation challenge \cite{cooke2010monaural} only contains a mixture of two speakers, with a limited vocabulary and insufficient training data.  The SISEC challenge (e.g., \cite{vincent2012signal}) is limited in size and designed for the evaluation of multi-channel separation, which can be easier than single-channel separation in general.

A training set consisting of 30 hours of two-speaker mixtures was generated by randomly selecting utterances by different speakers from the WSJ0 training set {\small \verb|si_tr_s|}, and by mixing them at various signal-to-noise ratios (SNR) between 0~dB and 5~dB.
We also designed the two training subsets from the above whole training set ({\it whole}), one considered the balance of the mixture of the genders ({\it balanced}, 22.5 hours), and the other only used the mixture of female speakers ({\it female}, 7.5 hours).
10 hours of cross validation set were generated similarly from the WSJ0 training set, which is used to optimize some tuning parameters, and to evaluate the source separation performance of the closed speaker experiments ({\bf closed speaker set}).
5 hours of evaluation data was generated similarly using utterances from sixteen speakers from the WSJ0 development set {\small \verb|si_dt_05|} and evaluation set {\small \verb|si_et_05|}, which are based on the different speakers from our training and closed speaker sets ({\bf open speaker set}). 
Note that many existing speech separation methods (e.g., \cite{smaragdis2007convolutive,LeRoux2015SparseNMF03}) cannot handle the  open speaker problem without special adaptation procedures, and generally require knowledge of the speakers in the evaluation.  
For the evaluation data, we also created 100 utterances of three-speaker mixtures for each closed and open speaker set as an advanced setup. All data were downsampled to 8~kHz before processing to reduce computational and memory costs.

The input features $X$ were the log short-time Fourier spectral magnitudes of the mixture speech, computed with a 32~ms window length, 8~ms window shift, and the square root of the hann window.   
To ensure the local coherency, the mixture speech was segmented with the length of 100 frames, roughly the length of one word in speech, and processed separately to output embedding $V$ based on the proposed model.
The ideal binary mask was used to build the target $Y$ when training our network. The ideal binary mask gives ownership of a time-frequency bin to the source whose magnitude is maximum among all sources in that bin. The mask values were assigned with 1 for active and 0 otherwise (binary), making $YY^T$ as the ideal affinity matrix for the mixture.
 
To avoid problems due to the silence regions during separation, a binary weight for each time-frequency bin was used during the training process, only retaining those bins such that each source's magnitude at that bin is greater than some ratio (arbitrarily set to -40 dB) of the source's maximum magnitude. 
Intuitively, this binary weight guides the neural network to ignore bins that are not important to all sources. 

\subsection{Training procedure}
\label{sec:train}
Networks in the proposed model were trained given the above input $X$ and the ideal affinity matrix $Y Y^T$.
The network structure used in our experiments has two bi-directional long short-term memory (BLSTM) layers, followed with one feedforward layer. Each BLSTM layer has 600 hidden cells and the feedforward layer corresponds with the embedding dimension (i.e., $K$). Stochastic gradient descent with momentum 0.9 and fixed learning rate $10^-5$ was used for training. In each updating step, a Gaussian noise with zero mean and 0.6 variance was added to the weight. 
We prepared several networks used in the speech separation experiments using different embedding dimensions from $5$ to $60$. 
In addition, two different activation functions (logistic and tanh) were explored to form the embedding $V$ with different ranges of $v_{n, k}$.
For each embedding dimension, the weights for the corresponding network were initialized randomly from the scratch according to a normal distribution with zero mean and 0.1 variance with the tanh activation and {\it whole} training set.
In the experiments of a different activation (logistic) and different training subsets ({\it balanced} and {\it female}), the network was initialized with the one with the tanh activation and {\it whole} training set. The implementation was based on CURRENNT, a publicly available training software for DNN and (B)LSTM networks with GPU support ({\small{\url{https://sourceforge.net/p/currennt}}}).

\subsection{Speech separation procedure}
In the test stage, the speech separation was performed by constructing a time-domain speech signal based on time-frequency masks for each speaker.
The time-frequency masks for each source speaker were obtained by clustering the row vectors of embedding $V$, where $V$ was outputted from the proposed model for each segment (100 frames), similarly to the training stage.
The number of clusters corresponds to the number of speakers in the mixture.
We evaluated various types of clustering methods: $k$-means on the whole utterance by concatenating the embeddings $V$ for all segments; $k$-means clustering within each segment; spectral clustering within each segment. For the within-segment clusterings, one needs to solve a permutation problem, as clusters are not guaranteed to be consistent across segments. For those cases, we report oracle permutation results (i.e., permutations that minimize the $L^2$ distance between the masked mixture and each source's complex spectrogram) as an upper bound on performance.

One interesting property of the proposed model is that it can potentially generalize to the case of three-speaker mixtures without changing the training procedure in Section \ref{sec:train}.
To verify this, speech separation experiments on three-speaker mixtures were conducted using the network trained with two speaker mixtures, simply changing the above clustering step from 2 to 3 clusters. Of course, training the network including mixtures involving more than two speakers should improve performance further, but we shall see that the method does surprisingly well even without retraining.

As a standard speech separation method, supervised sparse non-negative matrix factorization (SNMF) was used as a  baseline~\cite{LeRoux2015SparseNMF03}. While SNMF may stand a chance separating speakers in male-female mixtures when using a concatenation of bases trained separately on speech by other speakers of each gender, it would not make sense to use it in the case of same-gender mixtures. To give SNMF the best possible advantage, we use an oracle: at test time we give it the basis functions trained on the actual speaker in the mixture. For each speaker, 256 bases were learned on the clean training utterances of that speaker. Magnitude spectra with 8 consecutive frames of left context were used as input features. At test time, the basis functions for the two speakers in the test mixture were concatenated, and their corresponding activations computed on the mixture. The estimated models for each speaker were then used to build a Wiener-filter like mask applied to the mixture, and the corresponding signals reconstructed by inverse STFT.

For all the experiment, performance was evaluated in terms of averaged signal-to-distortion ratio (SDR)
using the \verb|bss_eval| toolbox \cite{vincent2006performance}. 
The initial SDR averaged over the mixtures was $0.16$ dB for two speaker mixtures and $-2.95$ dB for three speaker mixtures.

\begin{table}[t]
\caption{SDR improvements (in dB) for different clustering methods. }
\label{tab:K40}
\begin{center}
{\small
\begin{tabular}{lcc}
\hline 
\multicolumn{1}{l}{\bf method}  &\multicolumn{1}{c}{\bf closed speaker set}  &\multicolumn{1}{c}{\bf open speaker set } 
 \\
 \hline
oracle NMF  & 5.06 & - 
\\\hline
DC oracle $k$-means      &  6.54 & 6.45\\
DC oracle spectral       &  6.35 &6.26 \\
DC global $k$-means    & 5.87   &5.81 \\
\end{tabular}
}
\end{center}
\vspace{-0.7cm}
\end{table}

\begin{table}[t]
\caption{SDR improvements (in dB) for different embedding dimensions $K$ and activation functions}
\label{tab:diff_K}
\begin{center}
{\small
\begin{tabular}{lccccc}
\hline 
 &\multicolumn{2}{c}{\bf closed speaker set }  &\multicolumn{2}{c}{\bf open speaker set } \\
 \multicolumn{1}{l}{\bf model}  & DC oracle  &DC global & DC oracle  &DC global
\\ \hline      
$K=5$      &  -0.77 & -0.96 &  -0.74 & -1.07\\
$K=10$      &  5.15 &4.52  & 5.29 &4.64\\
$K=20$   &  6.25&5.56 &  6.38&5.69\\
$K=40$    & 6.54   &5.87 & 6.45  &5.81 \\
$K=60$   &6.00  &5.19 & 6.08 &5.28\\
 \hline  
$K=40$ logistic  & 6.59  &5.86  & 6.61 &5.95\\
\end{tabular}
}
\end{center}
\vspace{-0.5cm}
\end{table}

\begin{table}[t]
\caption{SDR improvement (in dB) for each type of mixture. Scores averaged over male-male (m+m), female-female (f+f), female-male (f+m), or all mixtures.}
\label{tab:gender}
\begin{center}
	{\small
\begin{tabular}{lllllllllllll}
\hline 
&  \multicolumn{1}{l} {\bf training gender } &	 \multicolumn{4}{c} {\bf closed speaker set }&	 \multicolumn{4}{c} {\bf open speaker set }\\
\multicolumn{1}{l}{\bf method}&  \multicolumn{1}{l} {\bf distribution }  &m+m &f+f  &f+m&all&m+m &f+f  &f+m&all\\
 \hline 
oracle NMF  		&  speaker dependent& 3.25& 3.31& 6.53 & 4.90&-&-&-&-\\
 \hline 
\multirow{3}{*}{\begin{tabular}{@{}c@{}}DC oracle \\ permute\end{tabular}} &  {\it whole}    & 3.79 & 4.29 & 9.04 &6.54&4.49&3.21&8.69&6.45\\
  	& {\it balanced} & 3.89 & 4.35 & 8.74 & 6.42&4.61&3.49&8.27&6.41\\
 	 	& {\it female}   & - & 5.03 & - & -&-&4.04&-&-\\
 \hline 
\multirow{3}{*}{\begin{tabular}{@{}c@{}}DC global \\ $k$-means\end{tabular}}	&   {\it whole} 	  & 2.54   & 2.85 &  9.07 & 5.87&3.51&1.42&8.57 &5.80\\
&   {\it balanced}	  & 2.78 & 2.87 & 8.63  & 5.72& 3.89&1.74&8.27&5.83\\
&    {\it female} 	  & - & 3.88 &  - & -&-&2.56&-&-
\end{tabular}
}
\end{center}
\vspace{-0.7cm}
\end{table}

\begin{table}[t]
\caption{SDR improvement (in dB) for three speaker mixture}
\label{tab:three_spk}
\begin{center}
	{\small
\begin{tabular}{lcc}
\hline 
\multicolumn{1}{l}{\bf method}  &\multicolumn{1}{c}{\bf closed speaker set }  &\multicolumn{1}{c}{\bf open speaker set }
\\ \hline 
oracle NMF   & 4.42 & -\\
DC oracle      &  3.50 & 2.81\\
DC global      & 2.74   & 2.22 \\

\end{tabular}
}
\end{center}
\vspace{-0.6cm}
\end{table}

\begin{figure}[ht]
	\centering
    \vspace{-.5cm}
	\includegraphics[width=0.88\columnwidth]{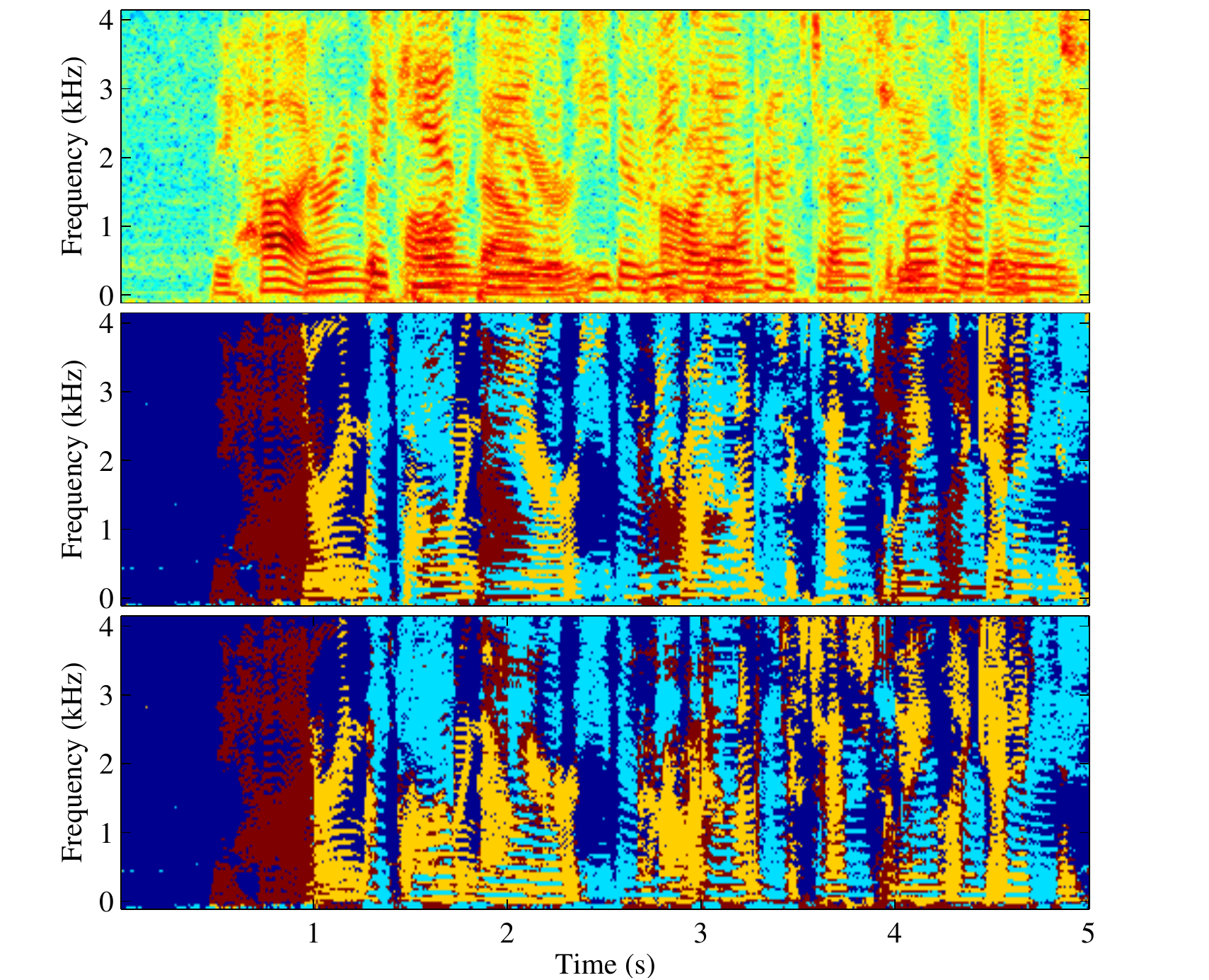}
		\vspace{-.3cm}
	\caption{An example of three-speaker separation. Top: log spectrogram of the input mixture. Middle: ideal binary mask for three speakers. The dark blue shows the silence part of the mixture.  Bottom: output mask from the proposed system trained on two-speaker mixtures.}
	\label{fig:3speakers}	\vspace{-.3cm}
\end{figure} 

\begin{figure}[ht]
	\centering
	\includegraphics[width=0.88\columnwidth]{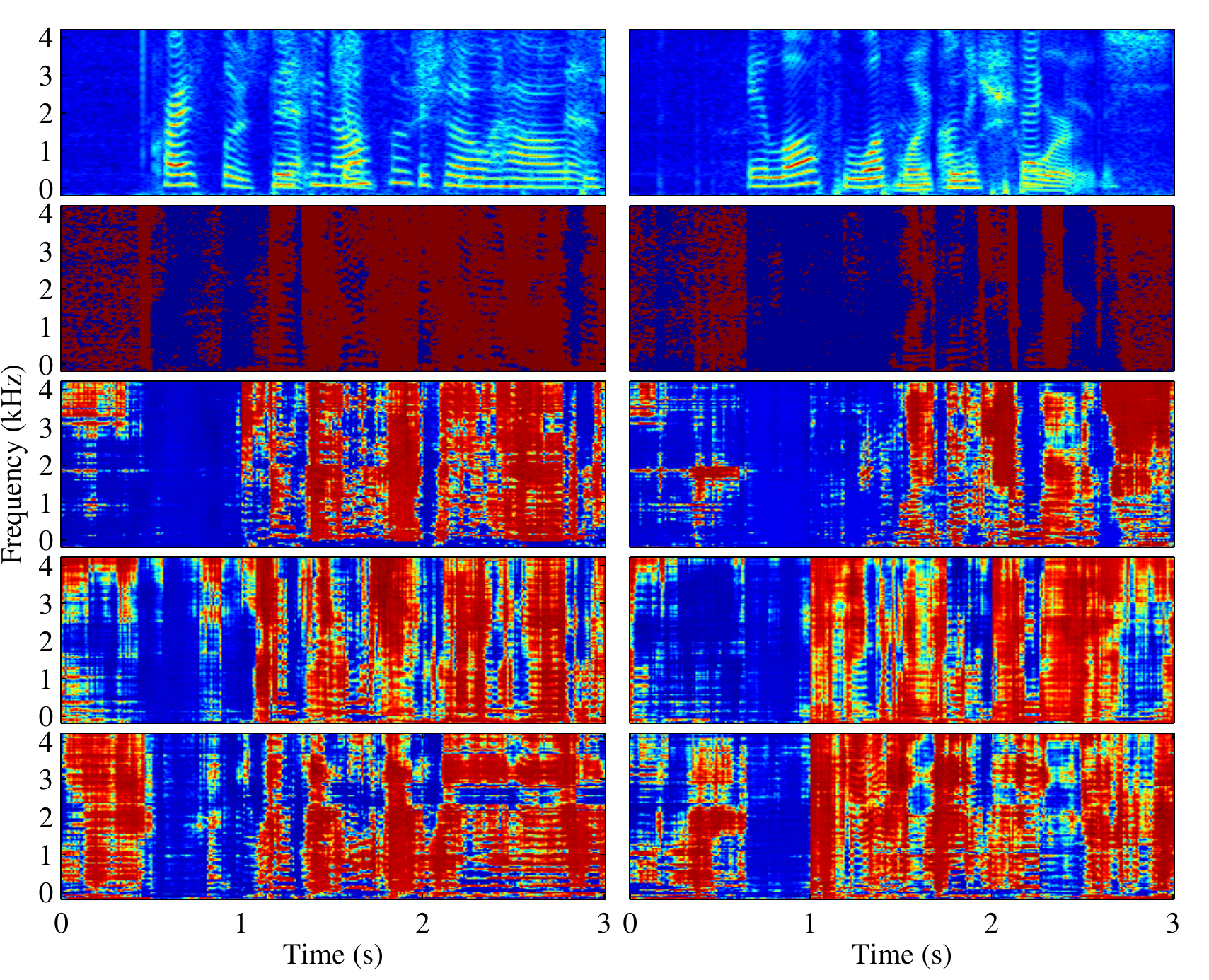}
	\vspace{-.3cm}
	\caption{Examples of embeddings for two mixtures:  f+f (left) and f+m (right). 1st row: spectrogram; 2nd row: ideal binary mask; 3rd-5th row: embeddings.}
	\label{fig:embed}	\vspace{-.5cm}
\end{figure}

\section{Results and discussion}

As shown in Table \ref{tab:K40}, both the oracle and non-oracle clustering methods for proposed system significantly outperform the oracle NMF baseline, even though the oracle NMF is a strong model with the important advantage of knowing the speaker identity and has speaker-dependent models. For the proposed system the open speaker performance is similar to the closed speaker results, indicating that the system can generalize well to unknown speakers, without any explicit adaptation methods. 
For different clustering methods, the oracle $k$-means outperforms the oracle "spectral clustering" by $0.19$ dB showing that the embedding represents centralized clusters.   To be fair, what we call spectral clustering here is using our outer product kernel instead of a local kernel function such as a Gaussian, as commonly used in spectral clustering.  However a Gaussian kernel could not be used here due to computational complexity.     
Also note that the oracle clustering method in our experiment resolves the permutation of two (or three in Table \ref{tab:three_spk}) speakers in each segment. In the dataset, each utterance usually contains 6$\sim$8 segments so the permutation search space is relatively small for each utterance.  Hence this problem may have an easy solution to be explored in future work. For the non-oracle experiments, the whole utterance clustering also performs relatively well compared  to baseline.  Given the fact that the system was only trained with individual segments, the effectiveness of the whole utterance clustering suggests that the network learns features that are globally important, such us pitch, timbre etc. 

In Table \ref{tab:diff_K}, the $K=5$ system completely fails, either because optimization of the current network architecture fails, or the embedding fundamentally requires more dimensions. The performance of $K=20$, $K=40$, $K=60$ are similar, showing that the system can operate in a wide range of parameter values.  
We arbitrarily used tanh networks in most of the experiments because the tanh network has larger embedding space than logistic network. However, in Table \ref{tab:diff_K}, we show that in retrospect the logistic network performs slightly better than the tanh one.  

In Table \ref{tab:gender}, since the female and male mixture is an intrinsically easier segmentation problem, the performance of mixture between female and male is significantly better than the same gender mixtures for all situations. As mentioned in Section \ref{sec:exp}, the random selection of speaker would also be a factor for the large gap. With more balanced training data, the system has better performance for the same gender separation with a sacrifice of its performance for different gender mixture. If we only focus on female mixtures, the performance is still better.  

Figure~\ref{fig:embed} shows an example of embeddings for two different mixtures (female-female and male-female), in which a few embedding dimensions are plotted for each time-frequency bin in order to show how they are sensitive to different aspects of each signal. 

In Table \ref{tab:three_spk}, the proposed system can also separate the mixture of three speakers, even though it is only trained on two-speaker mixtures. As discussed in previous sections, unlike many separation algorithms, deep clustering can naturally scale up to more sources, and thus make it suitable for many real world tasks when the number of sources is not available or fixed. Figure~\ref{fig:3speakers} shows one example of the separation for three speaker mixture in the open speaker set case.
Note that we also did experiments with mixtures of three fixed speakers for the training and testing data, and the SDR improvement of the proposed system is $6.15$.

Deep clustering has been evaluated in a variety of conditions and parameter regimes, on  a challenging speech separation problem.  Since these are just preliminary results, we expect that further refinement of the model will lead to significant improvement. For example, by combining the clustering step into the embedding BLSTM network using the deep unfolding technique\cite{Hershey2014arXiv09}, the separation could be jointly trained with embedding and lead to potential better result. Also in this work, the BLSTM network has a relatively uniform structure. Alternative architectures with different time and frequency dependencies, such as deep convolutional neural networks \cite{farabet2013learning}, or hierarchical recursive embedding networks \cite{sharma2014recursive}, could also be helpful in terms of learning and regularization.   Finally, scaling up training on databases of more disparate audio types, as well as applications to other domains such as image segmentation, would be prime candidates for future work.

\clearpage

\small{
	
\bibliographystyle{IEEEbib}
\bibliography{deep_clustering}
	
}

\end{document}